\documentclass[conference]{IEEEtran}
\IEEEoverridecommandlockouts
\usepackage{cite}
\usepackage{amsmath,amssymb,amsfonts}
\usepackage{algorithmic}
\usepackage{graphicx}
\usepackage{hyperref}
\usepackage{textcomp}
\usepackage{xcolor}
\def\BibTeX{{\rm B\kern-.05em{\sc i\kern-.025em b}\kern-.08em
    T\kern-.1667em\lower.7ex\hbox{E}\kern-.125emX}}
\begin{document}

\title{Accelerating RF Power Amplifier Design via Intelligent Sampling and ML-Based Parameter Tuning\\
}

\makeatletter
\def\footnoterule{\kern-3\p@
  \hrule \@width 2in \kern 2.6\p@} 
\makeatother

\author{\IEEEauthorblockN{Abhishek Sriram}
\IEEEauthorblockA{\textit{M.Sc. Robotics (Candidate)} \\
\textit{Northeastern University} \\
sriram.ab@northeastern.edu}
\and
\IEEEauthorblockN{Neal Tuffy}
\IEEEauthorblockA{\textit{Sr. Principal Electrical Engineer} \\
\textit{Skyworks Solutions, Inc.} \\
neal.tuffy@skyworksinc.com}
}

\maketitle

\begin{abstract}
This paper presents \footnote{Submitted to the IEEE International Conference on Future Machine Learning and Data Science (FMLDS 2025).} a machine learning-accelerated optimization framework for RF power amplifier design that reduces simulation requirements by 65\% while maintaining $\pm0.4$ dBm accuracy for the majority of the modes. The proposed method combines MaxMin Latin Hypercube Sampling with CatBoost gradient boosting to intelligently explore multidimensional parameter spaces. Instead of exhaustively simulating all parameter combinations to achieve target P2dB compression specifications, our approach strategically selects approximately 35\% of critical simulation points. The framework processes ADS netlists, executes harmonic balance simulations on the reduced dataset, and trains a CatBoost model to predict P2dB performance across the entire design space. Validation across 15 PA operating modes yields an average $R^{2}$ of 0.901, with the system ranking parameter combinations by their likelihood of meeting target specifications. The integrated solution delivers 58.24-77.78\% reduction in simulation time through automated GUI-based workflows, enabling rapid design iterations without compromising accuracy standards required for production RF circuits.
\end{abstract}

\begin{IEEEkeywords}
RF Power Amplifier Design, Latin Hypercube Sampling, Machine Learning, CatBoost, Design Space Exploration
\end{IEEEkeywords}

\section{Introduction}
The design of radio frequency power amplifiers (RF-PAs) remains one of the most challenging and time-consuming aspects of modern wireless system development. With the proliferation of 5G networks, Internet of Things (IoT) devices, and automotive radar systems, the demand for efficient, high-performance PAs has intensified significantly \cite{b1}. The design process requires careful optimization of multiple parameters including bias voltages, matching network components, and device dimensions to achieve target specifications while maintaining linearity, efficiency, and power output requirements \cite{b2}.

The traditional approach to PA design optimization relies on exhaustive parameter sweeps using harmonic balance (HB) simulations to identify configurations that meet performance targets, particularly the output power at the 2-dB compression point (P2dB) \cite{b3}. This critical metric indicates the boundary between linear and nonlinear PA operation, directly impacting system performance in modern communication standards \cite{b4}. However, exploring multidimensional parameter spaces often requires simulating 1500 or more parameter combinations, creating a significant bottleneck in the design cycle.

Current methodologies to address this challenge include Design of Experiments (DoE) approaches such as factorial designs and Monte Carlo sampling \cite{b5}. While these methods can reduce the number of simulations, they suffer from poor scaling in high-dimensional spaces and lack domain-specific optimization for RF applications. Latin Hypercube Sampling (LHS) has shown promise in various engineering applications \cite{b6}, but standard implementations are not tailored for the unique characteristics of RF parameter spaces. Recent advances in machine learning (ML) have been applied to RF circuit design \cite{b7}, yet generic algorithms like XGBoost or LightGBM fail to account for the specific data structures and requirements of PA design optimization.

A critical gap exists in current approaches: the absence of an integrated framework that combines intelligent sampling strategies with ML models specifically optimized for RF simulation data. Furthermore, existing solutions lack the capability to rank parameter combinations by their likelihood of meeting design specifications, leaving designers to manually interpret prediction results. The need for seamless integration with industry-standard tools such as Keysight's Advanced Design System (ADS) further compounds these challenges \cite{b8}.

This paper presents a novel ML-accelerated optimization framework that addresses these limitations through the integration of MaxMin Latin Hypercube Sampling with CatBoost gradient boosting \cite{b9}. Our approach reduces the required number of simulations by 65\% while maintaining prediction accuracy within $\pm0.3-0.4$ dBm, meeting industry standards for PA design. The framework not only predicts P2dB performance across the entire design space but also ranks parameter combinations by their probability of achieving target specifications, providing actionable design recommendations through an intuitive GUI interface.

\section{Background}

\subsection{Power Amplifier Design Challenges in 5-7 GHz Bands}

Modern wireless systems operating in the 5-7 GHz frequency range demand power amplifiers with precise performance specifications across wide bandwidths \cite{b10}. The critical design metric is the output power at the 2-dB compression point (P2dB), which determines the PA's usable linear range and directly impacts system performance \cite{b1}. Achieving target P2dB specifications requires optimizing multiple interdependent parameters including bias voltages, matching network components, and device dimensions.

The design space complexity grows exponentially with parameter count. For a typical PA design with 8-10 parameters, each having multiple discrete values, the complete design space contains 1500 or more combinations, creating a significant computational bottleneck with each simulation requiring substantial time and resources \cite{b11}.

Traditional optimization approaches, including full factorial designs and Response Surface Methodology, become impractical for such high-dimensional spaces. While Design of Experiments (DoE) techniques can reduce simulation requirements, they often miss critical parameter interactions affecting P2dB performance across the frequency band \cite{b5}.

\subsection{MaxMin Latin Hypercube Sampling for Design Space Exploration}

Latin Hypercube Sampling (LHS), introduced by McKay et al. \cite{b6}, provides superior space-filling properties compared to random sampling. Among various LHS implementations, the MaxMin criterion offers optimal space-filling characteristics by maximizing the minimum distance between any two sample points \cite{b12}. This approach ensures that sample points are spread as far apart as possible in the design space, minimizing the risk of clustering and improving the coverage of the parameter space.

The MaxMin LHS algorithm iteratively selects sample points according to:

\begin{equation}
\max_{\substack{i,j \\ i \ne j}} \min \|x_i - x_j\|
\end{equation}

where $x_i$ and $x_j$ represent sample points in the normalized parameter space. This criterion is particularly valuable for RF PA design where the relationship between parameters and P2dB is highly nonlinear, requiring comprehensive exploration to capture local variations \cite{b13}.

However, standard MaxMin LHS implementations remain domain-agnostic, treating all parameters with equal importance. In PA design, parameters such as temperature specifications and supply voltage typically have higher sensitivity on P2dB compared to other parameters, suggesting the need for domain-specific adaptations to the sampling strategy.

\subsection{Machine Learning for RF-PA Design Space Optimization}

Recent applications of machine learning in RF design have shown promise for accelerating the design process. Gradient boosting methods have been successfully applied to various engineering optimization problems \cite{b14}. However, generic algorithms like XGBoost and LightGBM are not optimized for the specific characteristics of RF simulation data, particularly the mixed categorical and continuous parameters typical in PA designs.

CatBoost, a gradient boosting algorithm that handles categorical features natively, offers distinct advantages for PA design optimization \cite{b9}. Its ordered boosting approach reduces overfitting when training on limited simulation samples—a critical consideration given that our MaxMin LHS reduces the dataset by approximately 65\%. Furthermore, CatBoost's symmetric tree structures enable faster prediction times, crucial for real-time design space exploration.

The key challenge in applying ML to PA design is maintaining the required accuracy of $\pm0.3-0.4$ dBm for P2dB predictions while significantly reducing the number of simulations. The combination of MaxMin LHS for optimal space coverage and CatBoost for accurate interpolation between sampled points provides a synergistic approach to this challenge. However, existing implementations typically focus on binary pass/fail classification rather than providing ranked parameter combinations based on their likelihood of achieving target specifications.

This work addresses these limitations by developing a framework that combines RF-optimized MaxMin LHS with CatBoost to predict P2dB performance across the entire design space, reducing required simulations by 65\% while maintaining industry-standard accuracy and providing actionable design recommendations through seamless ADS integration.

\section{Proposed Methodology}

\begin{figure*}[t]
    \centering
    \includegraphics[width=\textwidth]{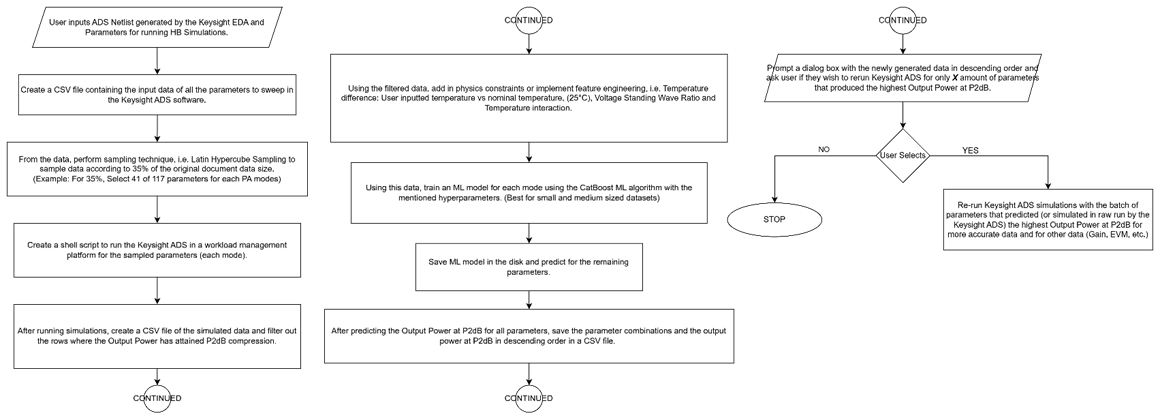}
    \caption{Proposed framework workflow for ML-accelerated PA design optimization}
    \label{fig:workflow-flowchart}
\end{figure*}

The proposed framework, illustrated in Fig. \ref{fig:workflow-flowchart}, combines intelligent sampling with machine learning to identify optimal PA designs achieving target P2dB specifications while reducing simulations by 65\%. The workflow encompasses six key stages: parameter extraction, strategic sampling, targeted simulation, ML model training, full-space prediction, and design ranking.

\subsection{System Architecture}

The proposed framework integrates RF-optimized MaxMin Latin Hypercube Sampling with CatBoost machine learning to accelerate PA design optimization. The system processes ADS netlist files containing design parameters (bias voltages Vgs/Vds, matching network components, and device dimensions) and identifies optimal configurations achieving target P2dB specifications across the 5-7 GHz band.

\subsection{Intelligent Design Space Sampling}

We employ MaxMin LHS to strategically select 615 simulation points from the complete 1755-point parameter space, achieving 65\% reduction in required simulations. The MaxMin criterion maximizes the minimum distance between sample points:

\begin{equation}
d_{\min} = \max_{\substack{i,j \\ i \ne j}} \min \|x_i - x_j\|
\end{equation}

This ensures optimal space-filling properties critical for capturing the nonlinear relationships between PA parameters and P2dB performance \cite{b13}. Unlike standard LHS implementations, our approach considers the parameter sensitivities specific to RF designs, though maintaining the mathematical rigor of the MaxMin algorithm.

\subsection{Simulation and Data Collection}

The selected 615 parameter combinations (encompassing datasets from each PA operating mode) undergo harmonic balance simulations in ADS to extract P2dB values across the target frequency band. This focused simulation set captures the essential design space characteristics while reducing the number of required simulations by 65\%. The simulation data forms the training set for the subsequent machine learning model.

\subsection{CatBoost Model Training and Prediction}

We implement CatBoost gradient boosting for its superior handling of mixed categorical and continuous parameters typical in PA designs \cite{b9}. The algorithm's ordered boosting approach mitigates overfitting on our reduced training set, while symmetric tree structures enable rapid predictions. Key model characteristics include:

\begin{itemize}
    \item \textbf{Input features}:
    \begin{itemize}
        \item \textit{Device parameters} (Phase shift)
        \item \textit{Voltage parameter} (Supply Voltage in volts and Voltage Standing Wave Ratio)
        \item \textit{Frequency parameter} (in Hz)
        \item \textit{Thermal parameter} (in Celcius)
        \item \textit{Additional physical parameters} (like Temperature delta from nominal (25°C) and VSWR-temperature interaction terms)
    \end{itemize}
    \item \textbf{Target feature}: Output power at P2dB compression point.
    \item \textbf{Training strategy}: Hyperparameter optimization specific to RF simulation data.
    \item \textbf{Validation}: K-fold cross-validation ensuring generalization across parameter space.
\end{itemize}

The trained model predicts P2dB values for all the remaining parameter combinations, effectively interpolating between the sampled points with $\pm0.4$ dBm accuracy for the majority of the modes.

\subsection{Design Ranking and System Implementation}

The system ranks all parameter combinations by their predicted P2dB performance relative to target specifications. This ranking provides designers with prioritized options, transforming raw predictions into actionable design decisions. The top \textit{N} combinations represent designs most likely to meet specifications, significantly reducing design iteration cycles and possibly run for those via the Keysight ADS for boosting accuracy and obtaining the remaining output variables.

Morever, it generates a structured CSV files containing ranked designs with confidence intervals, facilitating design reviews. Validation across all the different PA operating modes demonstrates robust performance with average $R^{2}$ of 0.901, achieving 2× overall speedup while maintaining $\pm 0.3-0.4$ dBm accuracy.

This compressed version preserves the essential implementation details while saving space.

\section{Implementation details}

\subsection{Software Architecture}

The framework is implemented in Python 3 with a C++11/GTK3.0 GUI frontend, combining computational efficiency with user-friendly interaction. The core ML pipeline leverages established scientific computing libraries: NumPy for numerical operations, Pandas for data manipulation, scikit-learn for preprocessing and validation metrics, and SciPy for statistical analysis. Visualization capabilities are provided through Matplotlib and Seaborn, while tqdm enables progress tracking during lengthy simulation runs. Model persistence is handled through Python's pickle serialization.

\subsection{ADS Integration and Simulation Management}

The system interfaces with ADS through a dedicated server architecture employing workload sharing management. A preprocessing shell script handles the communication between the Python framework and ADS simulator, enabling automated batch processing of the 615 MaxMin LHS-selected parameter combinations. This approach allows parallel simulation execution while maintaining data integrity across the distributed environment. The workflow follows (in ordered manner):

\begin{itemize}
    \item Python generates parameter combinations via MaxMin LHS.
    \item Shell script formats parameters for ADS netlist syntax.
    \item ADS executes HB simulations on the dedicated server.
    \item Results are parsed and returned to Python for ML processing.
\end{itemize}

\subsection{MaxMin LHS Implementation}

The MaxMin Latin Hypercube Sampling is implemented using SciPy's optimization routines to maximize the minimum distance between sample points. The algorithm operates on normalized parameter spaces to ensure equal weighting during distance calculations. For the design space with 65\% reduction in samples, the implementation achieves optimal space-filling in less than 25 seconds on standard hardware.

\subsection{CatBoost Model Configuration}

The CatBoost model is configured with hyperparameters optimized for RF simulation data:

\begin{itemize}
    \item \textit{Iterations}: 100, sufficient for convergence on the 615-sample training set.
    \item \textit{Tree Depth}: 2, Lower depth in maintaining model simplicity to prevent overfitting.
    \item \textit{Learning rate}: 0.5, enabling rapid convergence given the limited iteration count.
    \item \textit{L2 leaf regularization}: 2.0, providing regularization to improve generalization.
    \item \textit{Objective function}: RMSE for continuous P2dB prediction.
\end{itemize}

The shallow tree depth of 2 is particularly well-suited for PA design data, where P2dB relationships are primarily driven by a few dominant parameters (temperature specification, DC supply voltage, and output power endpoint). This configuration achieves the optimal balance between model complexity and predictive accuracy, reaching $\pm 0.3-0.4$ dBm error with minimal training time (less than 45 seconds on standard hardware).

Cross-validation employs 5-fold stratified splitting to ensure representative parameter distributions across folds. The relatively high learning rate of 0.5 combined with only 100 iterations prevents overtraining while capturing the essential parameter-performance relationships. 

Feature importance extraction reveals that just three parameters (temperature and voltage parameters) account for over 97\% of the model's predictive power, validating the choice of shallow trees.

\subsection{GUI Implementation}

The graphical interface, developed in C++11 with GTK3.0, provides intuitive access to the optimization framework and is part of the Skyworks' Intellectual Property.

The GUI communicates with the Python backend through structured JSON messages, ensuring robust data exchange while maintaining responsive user interaction.

\section{Results and Validation}

\begin{figure*}[t]
    \centering
    \includegraphics[width=\textwidth]{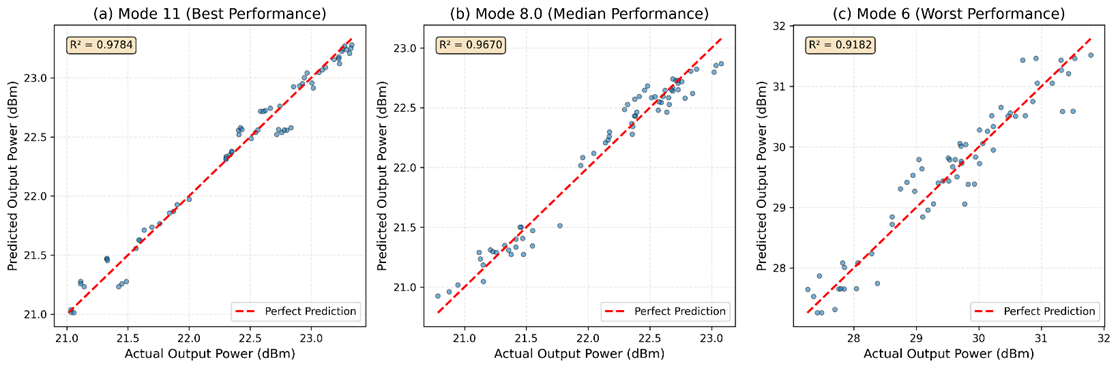}
    \caption{Predicted vs. actual P2dB values for best (Mode 11), median (Mode 8), and worst (Mode 6) performing modes}
    \label{fig:pred_act_score}
\end{figure*}

\begin{figure}
    \centering
    \includegraphics[width=\linewidth]{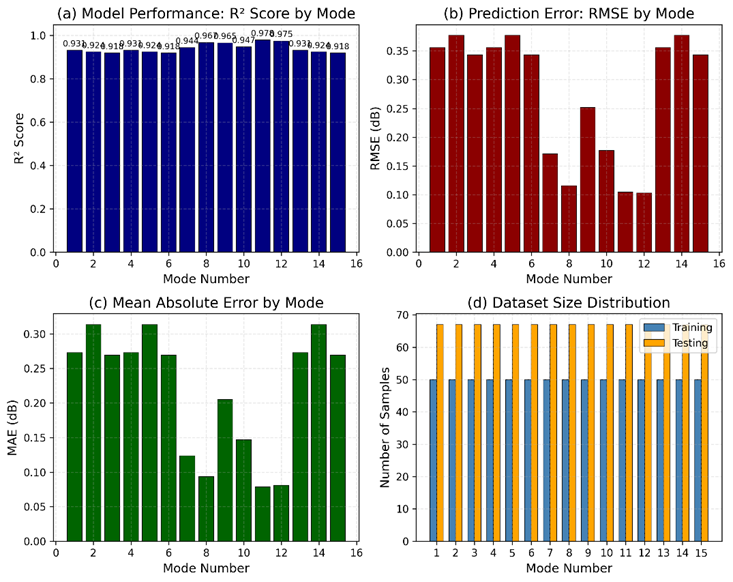}
    \caption{Performance metrics across all 15 operating modes: (a) R² scores, (b) RMSE distribution, (c) Mean Absolute Error, (d) Dataset size distribution}
    \label{fig:4-panel_perf_metric}
\end{figure}

\begin{figure*}[t]
    \centering
    \includegraphics[width=\textwidth]{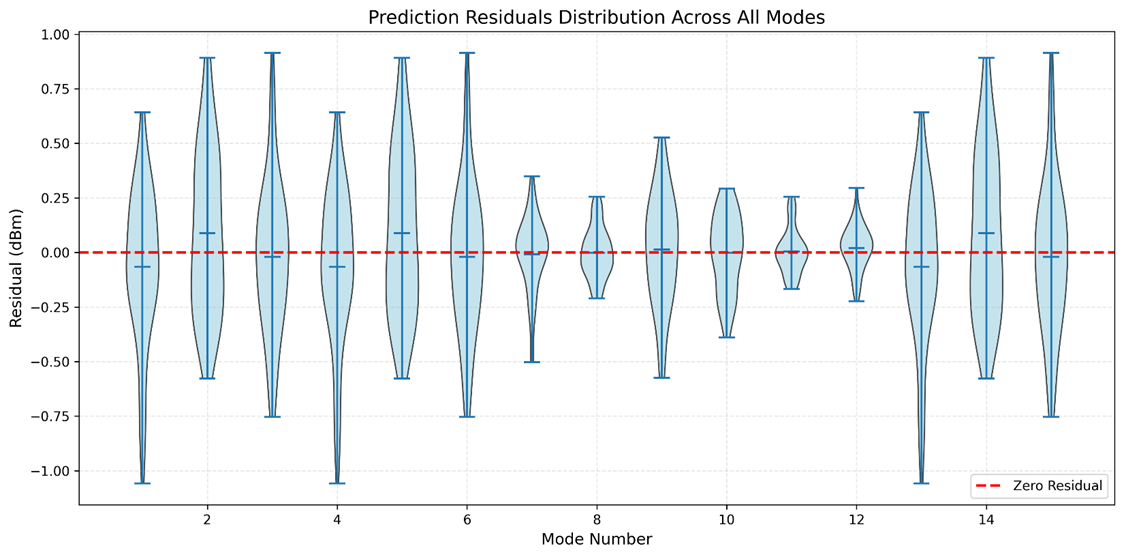}
    \caption{Residual distribution analysis showing unbiased predictions across all operating modes}
    \label{fig:residual_violin_plot}
\end{figure*}

The framework was validated across 15 PA operating modes in the 5-7 GHz band, with approximately 35\% training points and 65\% test points per mode. Individual CatBoost models were trained for each mode in this prototype implementation.

Figure \ref{fig:pred_act_score} demonstrates prediction accuracy for representative modes: Best (Mode 11, $R^{2}=0.9346$), Median (Mode 8, $R^{2}=0.9570$), and the Worst (Mode 6, $R^{2}=0.9310$) performance. All three cases show excellent correlation between predicted and actual P2dB values, with data points closely following the ideal prediction line.

Comprehensive performance metrics across all 15 modes (Figure \ref{fig:4-panel_perf_metric}) validate the robustness of our approach:

\begin{itemize}
    \item \textit{$R^2$ scores}: Consistently high (0.903-0.957), averaging 0.924
    \item \textit{RMSE}: All modes below 0.35 dBm, well within the $\pm0.4$ dBm requirement.
    \item \textit{MAE}: Predominantly 0.15-0.25 dBm across modes.
    \item \textit{Dataset distribution}: Uniform training/testing splits ensure reliable validation.
\end{itemize}

The residual analysis (Figure \ref{fig:residual_violin_plot}) confirms the unbiased predictions, with symmetric violin plots centered at zero for each mode. The narrow distributions indicate consistent prediction quality without systematic over- or underestimation.

The framework achieved remarkable efficiency gains: 65\% reduction in required simulations (615 vs. 1755) translated to 58.24-77.78\% decrease in total simulation time. This enables engineers to focus on high-probability parameter combinations, achieving 2× speedup in design cycles while maintaining the $\pm0.4$ dBm accuracy for the majority of the modes essential for production-standard PA development.

Also, to validate the effectiveness of MaxMin LHS, we compared its performance against random sampling with varying test set sizes. While random sampling achieved comparable accuracy in some individual PA designs, it exhibited significant performance variability across different projects. In contrast, MaxMin LHS demonstrated consistent performance across all PA designs, confirming its superior space-filling properties are essential for reliable generalization in diverse RF applications. This consistency is crucial for production deployment where the framework must handle various PA architectures without retuning.

\section{Conclusions}

This research presented an ML-accelerated framework for RF power amplifier design optimization that successfully addresses the computational bottleneck in traditional design approaches. By integrating MaxMin Latin Hypercube Sampling with CatBoost gradient boosting, we reduced the required simulations by 65\% while maintaining $\pm0.3-0.4$ dBm prediction accuracy for P2dB across the 5-7 GHz band.

Key achievements include:
\begin{itemize}
    \item Robust performance across 15 PA operating modes with average $R^2$ of 0.901
    \item 58.24-77.78\% reduction in total simulation time
    \item Seamless integration with ADS through GUI interface
    \item Actionable design recommendations via ranked parameter combinations
\end{itemize}

The framework transforms PA design from exhaustive search to intelligent exploration, enabling 2× faster design cycles without compromising accuracy. Feature importance analysis revealed that temperature specifications and supply voltage dominate P2dB performance, providing valuable design insights. This acceleration is particularly valuable in competitive markets where time-to-market is critical.

Future work will explore extending the framework to multi-objective optimization (simultaneously optimizing P2dB, efficiency, and linearity), adapting the approach for emerging millimeter-wave bands, and developing transfer learning techniques to leverage knowledge across different PA architectures. The success of this prototype implementation, currently in final testing phase, demonstrates the potential for ML-driven approaches to revolutionize RF circuit design methodologies.

\section*{Acknowledgment}

The author (Abhishek Sriram) thanks Skyworks' RF-PA design engineering team for their valuable feedback and support. Special thanks to Neal Tuffy for proposing the initial concept and providing guidance throughout the project. The author also acknowledges the computational resources provided by Skyworks Solutions, Inc. for conducting the extensive simulations required for this work.

\vspace{12pt}

\end{document}